\documentclass[sigconf]{acmart}

\usepackage{cite}
\usepackage{amsmath,amsfonts}
\usepackage{algorithmic}
\usepackage{graphicx}
\usepackage{textcomp}
\usepackage{subfig}

\copyrightyear{2020}
\acmYear{2020}
\setcopyright{acmlicensed}\acmConference[FDG '20]{International Conference on the Foundations of Digital Games}{September 15--18, 2020}{Bugibba, Malta}
\acmBooktitle{International Conference on the Foundations of Digital Games (FDG '20), September 15--18, 2020, Bugibba, Malta}
\acmPrice{15.00}
\acmDOI{10.1145/3402942.3402943}
\acmISBN{978-1-4503-8807-8/20/09}

\begin{document}

\title{Collaborative Agent Gameplay in the \emph{Pandemic} Board Game}
%\author{Anonymous}
%\begin{comment}
\author{Konstantinos Sfikas}
\affiliation{%
  \institution{Institute of Digital Games, University of Malta}
  \city{Msida}
  \country{Malta}
}
\email{konstantinos.sfikas@um.edu.mt}

\author{Antonios Liapis}
\affiliation{%
  \institution{Institute of Digital Games, University of Malta}
  \city{Msida}
  \country{Malta}
}
\email{antonios.liapis@um.edu.mt}
%\end{comment}

\begin{abstract}
While artificial intelligence has been applied to control players' decisions in board games for over half a century, little attention is given to games with no player competition. \emph{Pandemic} is an exemplar collaborative board game where all players coordinate to overcome challenges posed by events occurring during the game's progression. This paper proposes an artificial agent which controls all players' actions and balances chances of winning versus risk of losing in this highly stochastic environment. The agent applies a Rolling Horizon Evolutionary Algorithm on an abstraction of the game-state that lowers the branching factor and simulates the game's stochasticity. Results show that the proposed algorithm can find winning strategies more consistently in different games of varying difficulty. The impact of a number of state evaluation metrics is explored, balancing between optimistic strategies that favor winning and pessimistic strategies that guard against losing.
\end{abstract}




\maketitle

\section{Introduction}\label{sec:introduction}

Board games have fascinated researchers in Artificial Intelligence (AI) from the very beginnings of the field. Chess, checkers, and tic-tac-toe were some of the first testbeds \citep{yannakakis2018artificial} for AI algorithms such as reinforcement learning \citep{Samuel1959Checkers}. Events where a master player competed against a computer in a game of chess \citep{Newborn1997KasparovVD} and Go \citep{Gershgorn2016Alphago} garnered massive public interest. Perhaps due to the physical aspect of the board games, or their popularity, board game playing is still the most common way in which the general public perceives AI. Academic research in board game playing AI has of course moved beyond most pedestrian board games, applying a diverse set of algorithms for playing card games with millions of card combinations such as \emph{Magic: the Gathering} (Wizards of the Coast, 1993) \citep{Cowling2012EnsembleDI}, games of tactical card placement such as \emph{Lords of War} (Black Box, 2012) \citep{Sephton2014HeuristicMP} and \emph{Carcassonne} (Hans im Gl\"{u}ck, 2000) \citep{Heyden2009Carcassone}, card games of team-based competition such as \emph{Hanabi} (Abacusspiele, 2010) \citep{WaltonRivers2017EvaluatingAM} or \emph{Codenames} (Czech Games Edition, 2015) \citep{Summerville2019CodenamesAI}, and many more. 

Traditional board games such as chess \citep{Newborn1997KasparovVD} and backgammon \citep{tesauro1995tdgammon}, as well as recent card games such as \emph{Race for the Galaxy} (Rio Grande, 2007) \citep{duringer2017race} or digitized board games such as \emph{Hearthstone} (Blizzard, 2014) \citep{santos2017mctshearthstone,hoover2019hearthstone}, focus on players competing to deplete another player's resources (pawns, hit points) or to accumulate more victory points before the game ends. However, today's ecosystem of board games has a plethora of alternative modes of gameplaying. A particularly interesting type of board game invites \emph{collaborative} play, where all players must work together to survive (and win) against a rule-based system which presents an escalating challenge. Common design patterns for such collaborative games are (a) player roles specializing in certain tasks, (b) a rule-based system with high stochasticity (via drawn cards or dice) that introduces more and more complications and challenges to the game state, (c) a race against time for players to achieve victory, and (d) a dilemma between performing actions that mitigate current threats and actions that lead to victory. An example collaborative board game is \emph{Forbidden Island} (Gamewright, 2010) where the terrain tiles that make up the board may `submerge' and then be removed completely, based on a shuffled deck that determines which tile is affected. This pressures players to either save the tiles to increase their movement options (and avoid losing) or to pursue the winning criterion of collecting artefacts scattered across the board. Players choose roles which have increased mobility options or ignore/modify preconditions for some actions. More complex collaborative games such as \emph{Arkham Horror} (Fantasy Flight, 2005) and \emph{Robinson Crusoe} (Portal, 2012) follow similar patterns. \emph{Pandemic} (Z-Man, 2008) is one of the most popular collaborative board games and is fairly straightforward to play: players take different specialized roles and strive to cure diseases while these diseases infect more and more cities on the board with disease cubes. Players must balance between removing disease cubes (to stop the game from ending) while also exchanging cards in order to cure diseases (to win the game). What makes \emph{Pandemic} particularly interesting is that the cities that are infected are not chosen completely randomly; \emph{Pandemic} implements a clever system of recycling past infected cities. This means that players can anticipate the next few cities that will be infected (but not the order in which they will be infected) and strategize how best to minimize the risk.

This paper highlights that collaborative board game play poses its own set of challenges to Artificial Intelligence. While competitive play challenges AI to anticipate what the other player might do or how best to block another player, collaborative board game play challenges AI to best coordinate with other players. When all players are controlled by AI, a plan can be formulated for every player (by a single controller) and executed to the letter. This is actually how human players also handle a collaborative board game by making a strategy for every player's move and executing it (or replan, if circumstances change). The AI challenge of collaborative board game play is thus not to align each player's goal (as a single controller can control every player) but instead (a) to balance between short-term damage control and long-term strategies that win the game, (b) to optimally take advantage of different players' roles and special abilities, and (c) to anticipate the best- and worst-case scenarios of upcoming events and how they will affect the game state. Due to the stochastic nature of escalating threats posed by the game system, human players similarly perform risk assessment and mitigation in the hopes of avoiding the worst outcomes which usually cause the game to be lost.

The board game \emph{Pandemic} is chosen to test how collaborative board game play can be handled by AI. The controlled way in which the infected cities are reshuffled and may re-appear again and again makes for an interesting challenge for AI to assess risks of outbreaks. Failing to downsize the disease cubes or the number of outbreaks can lose the game; however, failing to cure all diseases swiftly also causes the game to be lost. The multiple ways in which the game can be lost, versus a single way in which it can be won, emphasizes the tension between pessimistic strategies (curbing losing conditions) and optimistic strategies (getting closer to a winning condition) within an AI agent's game state evaluation. Finally, the numerous actions available per player, combined with special abilities of each player role which modify these actions, requires an abstraction of the action space in order for AI to take only meaningful decisions on a more macro-strategic level. This paper takes first steps to address each of these challenges in an implementation of a Rolling Horizon Evolutionary Algorithm (RHEA) \citep{Liebana2013RollingHE} for controlling all players in a \emph{Pandemic} game session. The paper introduces a way of abstracting actions into more meaningful macro-actions, a forward model which accounts for the unknown distribution of the decks while capturing the probabilities of each threat, and necessary modifications to RHEA in order to account for different players' turn order and the need for an initial seed. The paper tests the RHEA controller in ten different testbed setups, and explores the impact of state evaluations which reward winning or penalize losing in different ways.

\begin{figure*}[t]
\centering
\includegraphics[trim=0cm 0cm 0cm 4cm,clip=true,width=0.95\textwidth]{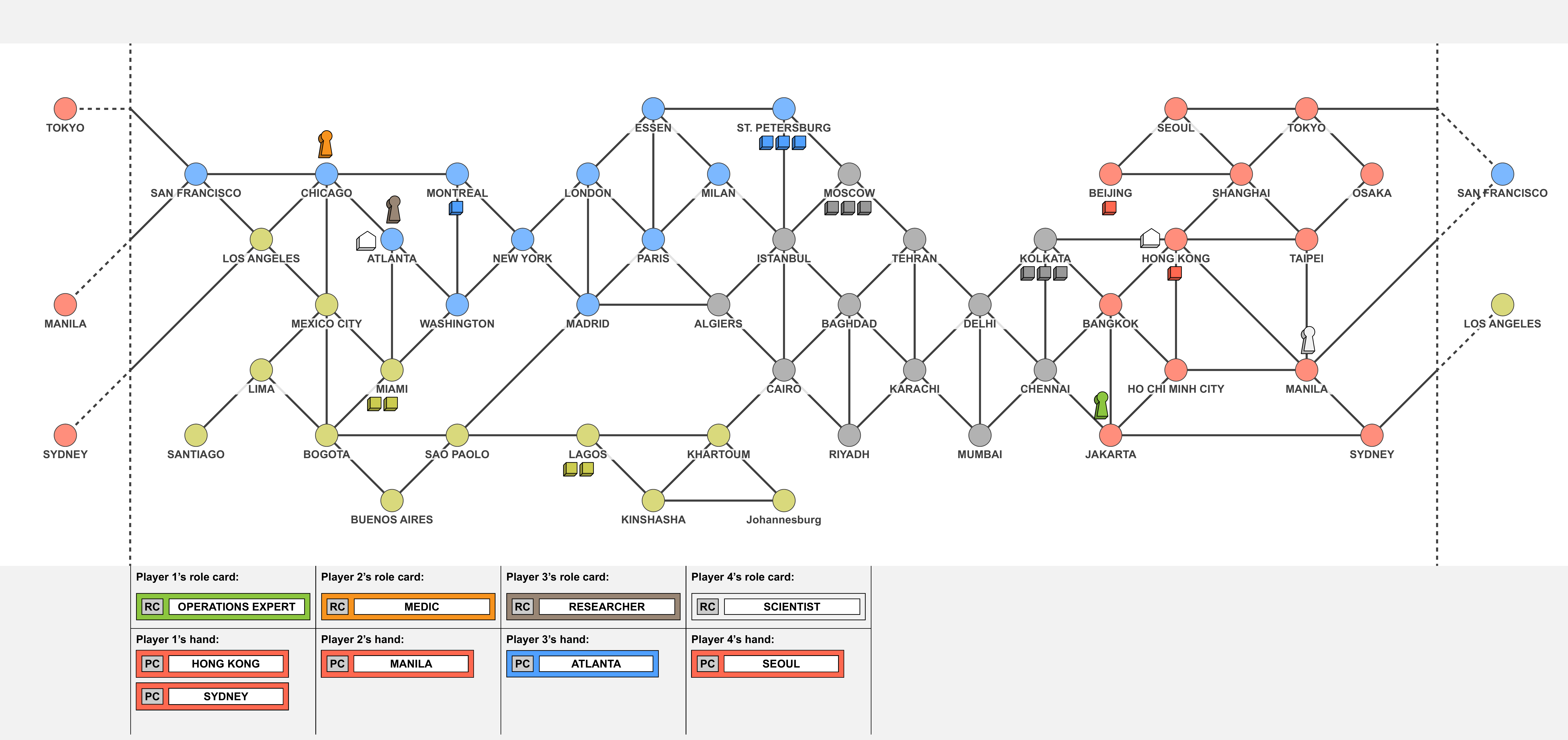}
\caption{Example game state during a playthrough of \emph{Pandemic}. Players take different roles (see bottom of the image), must travel the world to treat diseases (disease cubes on cities) and must use cards in their hand (under each player's role) to cure diseases at research stations (white houses on cities). The current game state is at the start of Player 2's turn.}
\label{fig:example}
\end{figure*}

\section{Related Work}\label{sec:related}

While AI research on board games has explored a vast range of algorithms, the majority of AI for board game play focuses on some form of game tree search. Early experiments in zero-sum games such as chess relied on the minimax algorithm \citep{turing1953games}, where the AI attempted to minimize losses from the opponents' expected (optimal) move. In complex games, each unique game state is difficult to enumerate, and the end condition is reached after many rounds of player actions. In such cases traditional tree search methods must be enhanced (a) by abstracting the current game state via carefully designed rule-based systems and via learned models \citep{Silver2016alphago}, or (b) by exploring only a small sample of future states.
Taking advantage of both strategies, Monte Carlo Tree Search has been especially powerful for board game play. Monte Carlo Tree Search (MCTS) builds a game tree in an incremental and asymmetric manner using a tree policy which balances exploration (sampling many strategies) and exploitation (expanding on more promising strategies) \citep{Browne2012ASO}. When the most urgent node of the tree is identified via the tree policy, a simulation (playout) from that node is performed using a \emph{default policy} which takes decisions on the agent's moves. The playout's end-state is evaluated and back-propagated through the selected nodes of the tree. Playouts may last until the game is won or lost, or until a maximum number of actions are taken. The default policy in playouts may be completely random (aheuristic) or take advantage of domain knowledge; the tradeoff is generality versus computational efficiency, respectively \citep{Drake2007heavy,Browne2012ASO}. By building partial, shallow trees and only assessing the terminal state after (inexact) playouts, MCTS is able to provide a valid next action \emph{anytime} \citep{Liebana2013RollingHE}, unlike other tree search algorithms such as A*. MCTS has shown very good results in deterministic board games \citep{Soemers2019BiasingMW} and can perform well in unknown games, e.g. in the General Video Game AI competition \citep{Park2015MCTSWI}. MCTS hinges on a forward model for simulating the game; when the game state changes stochastically due to the agent's actions or other factors, its performance can suffer \citep{Liebana2016AnalyzingTR}. While players' actions are deterministic in \emph{Pandemic}, the game state changes after each player's turn in unpredictable ways and MCTS is ineffective for handling this non-determinism.

The Rolling Horizon Evolutionary Algorithm (RHEA) was initially proposed in \citep{Liebana2013RollingHE} as a potent alternative of MCTS for real-time agent control problems \citep{Tong2019EnhancingRH}. RHEA evolves a sequence of actions in order to maximize some quality of the game state at the end of these actions, then performs only the first action of the fittest individual. As its name suggests, the performance of the RHEA hinges on its planning horizon $H$. Experiments in RHEA variants for the General Video Game AI competition \citep{Gaina2017RollingHE} highlighted the impact of the population size and length of the chromosome (horizon $H$) on performance. Similar to MCTS, the sequence of actions that make up the chromosome in RHEA are simulated via a forward model which can return the end-state after all actions are taken. Since RHEA does not require any part of a tree to be built, it can be more efficient than MCTS---especially in `noisy' environments where the forward model may falter. Since RHEA is limited to look-ahead only up to its horizon $H$, a state evaluation (fitness) of the game state after simulating all actions in the chromosome is vital. Unlike MCTS, which can in theory perform a random simulation until the game is won or lost, assessing an intermediate state of the game greatly affects the performance of RHEA.
Finally, due to the strong influence of the horizon $H$ on RHEA efficiency, a compact representation of the action space is preferred. Already in the first implementation of RHEA for games \citep{Liebana2013RollingHE}, the chromosome contained \emph{macro-actions} which could capture more substantial changes to the game state; in that real-time control problem, a \emph{macro-action} was the same action repeated 10 times. 

Of particular interest is the RHEA controller \citep{Bravi2019splendor} developed for \emph{Splendor} (Space Cowboys, 2014), a competitive card game where the stochasticity of cards appearing on the marketplace can severely affect which actions are available (or optimal) in future turns. \emph{Splendor} has similar patterns to \emph{Pandemic} which can challenge AI and forward models in particular, specifically the long-term implications of early-game actions, the limited turns before the game ends, and the stochasticity of hidden decks. Differences between the two include the competitive nature of \emph{Splendor} which necessitates opponent modeling \citep{Bravi2019splendor}; most importantly, stochasticity in \emph{Splendor} has far less impact than in \emph{Pandemic} since 12 marketplace cards are visible at all times, one unseen marketplace card may appear per turn, and marketplace cards in the same stack have fairly similar and always positive uses. In terms of AI control, \citep{Bravi2019splendor} dealt with the stochasticity of the forward model by implementing a random action generator which only produced valid actions based on the state, and used its seed as representation for RHEA. The algorithm described in this paper similarly handles stochasticity by operating on a much more constrained space of valid moves based on a hierarchical random action selection, but uses a constrained hierarchical mutator for actions, compared to the agnostic way in which mutation operates on random actions' seeds in \citep{Bravi2019splendor}. 

\section{The Game}\label{sec:game}
As noted in Section \ref{sec:introduction}, the \emph{Pandemic} board game is an ideal instance of collaborative gameplay. This section describes the game rules of \emph{Pandemic} and lists the nuances of the version tested in this paper.

\subsection{Components}\label{sec:game_components}
\emph{Pandemic} is played on a world map with 48 cities connected as a graph (see Fig.~\ref{fig:example}). Each city has one color (blue, yellow, red or black) and will be infected by disease cubes of that color during game setup and as the game progresses. The game has 24 disease cubes of each color, and a city can have a maximum of 3 cubes of the same color; if a fourth cube should be added, an outbreak occurs instead and all adjacent cities receive a cube of that color. Where disease cubes are added is determined by 48 infection cards, which match the 48 cities on the board. Each player has a pawn and can move from city to city and take actions (see Section \ref{sec:game_actions}). Finally, \emph{Pandemic} has a set of player cards and a set of epidemic cards which are shuffled together to form the player deck. A player card shows a city and its respective color and is drawn and kept by the acting player, while an epidemic card is not kept but influences the infection process as discussed in Section \ref{sec:game_state}.

\subsection{Player Actions}\label{sec:game_actions}
Each player has a hand of \emph{player cards}, and each card refers to a specific city on the board and its color. Players take turns acting, and at the end of a player's turn she gains two player cards and more cities of the board become infected (see Section \ref{sec:game_state}). On a player's turn, she can perform up to four actions: the possible action types include 4 ways of moving between cities, and 4 ways of changing the game state. Movement actions can (a) move the player along an edge from one city to an adjacent city (drive/ferry), (b) move the player from a city with a research station to any other city with a research station (shuttle flight). The other two movement actions require the player to discard a card, either to (c) move to the city on the discarded card from anywhere on the board (direct flight) or (d) to move to any city on the board if they are currently at the city on the discarded card (charter flight). Movement actions allow players to reach cities where they can perform the other action types available to them: \emph{treat disease} by removing one disease cube from a city they are currently in, \emph{build research station} at the city they are in by discarding a card with the same city, \emph{share knowledge} by giving or receiving a player card from a player on the same city, provided that the card traded also refers to the same city, or \emph{discover a cure} by discarding five cards of the same color at a city with a research station. Players can remove all disease cubes of a cured disease in the same city with a single treat action. Moreover, if no cubes of a cured disease exist on the board then it is eradicated and drawing cities of this color from the infection pile has no effect.

It should be noted that in the \emph{Pandemic} board game, players take different roles which modify some of the above actions, while special event cards are shuffled into the player deck which can then be used by players anytime. To simplify the AI controller's range of options, no special event cards are used in this version of \emph{Pandemic} while players can only choose the following player roles:
\begin{itemize}
\item \textbf{Operations Expert:} Can build a research station without discarding a city card, and can move from a research station to any city by discarding any city card (once per turn).
\item \textbf{Researcher:} Can give any city card from her hand to another player in the same city as her, without this card having to match her city.
\item \textbf{Medic:} Can remove all cubes of the same color with one treat disease action (or freely if the disease is cured). 
\item \textbf{Scientist:} Can discover a cure with 4 (not 5) player cards of the same color.
\end{itemize}

\subsection{Game State \& Ending Conditions}\label{sec:game_state}
In the beginning of the game, each player receives two player cards. The remaining player cards are split into piles of equal size and one epidemic is shuffled into each pile. In an Easy game, four epidemics are shuffled into four piles of player cards and then the different piles are placed one on top of each other. Finally, the infection deck is shuffled and 9 cities are revealed: the first 3 cities each receive 3 disease cubes of their respective colors, the next 3 cities receive 2 cubes each, and the last 3 cities receive 1 cube each. Revealed infection cards are placed on the infection discard pile.

Players take turns to perform 4 actions (see Section \ref{sec:game_actions}), and then receive two cards from the player deck. If they reveal city cards then they keep them in their hand, although they must discard any cards over 7 from their hand after drawing. If they reveal an epidemic card, the infection rate increases, the city at the bottom of the infection deck gains 3 matching cubes and is discarded; finally, all cards in the infection discard pile are shuffled and added (face down) on top of the infection deck. This mechanism ensures that cities that have already been infected will be soon infected again. After the players have drawn two player cards (and resolved any epidemic), a number of cards are drawn from the infection deck and one cube of matching color is added to each. The number of infection cards drawn depends on the infection rate (2, 3 or 4 infections for 0-3 epidemics, 4-5 epidemics, 6-7 epidemics respectively).

The game can only be won if players discover cures for all four diseases (through the discover a cure action in Section \ref{sec:game_actions}). The game can be lost if players need to place disease cubes of a certain color but no cubes of that color remain off the board, if the number of outbreaks reaches 8, or if the player deck runs out. This forces players to rush to discover a cure while controlling the cubes on the board so that the other losing conditions are not met.

\section{Rolling Horizon \emph{Pandemic} Agent}\label{sec:agent}

While players can easily identify a set of promising actions, the high branching factor and long-term repercussions of some player actions raise a serious challenge for game-playing AI. To make the problem easier to handle, a number of steps are taken to simplify playouts, and the RHEA was adapted to handle different players' turns. This section discusses the final RHEA architecture.

\subsection{Game Abstraction}\label{sec:abstraction}

\subsubsection{Forward Model}\label{sec:abstraction_forward}
A forward model is necessary in order to simulate the game and evaluate future states based on current actions. On the one hand, \emph{Pandemic} is highly stochastic as both the player deck and the infection deck are unknown to players. On the other hand, the stochasticity is somewhat known to players (especially halfway into the game), since the epidemics on the player deck are distributed fairly evenly in the beginning, and after the first epidemic the infection deck always has previously seen cities on top. 
To maintain this distribution, the forward model randomizes its infection deck and player deck as follows. For the infection deck, cards reinserted due to an epidemic are stored in separate stacks; each stack is shuffled on its own, and the infection deck is recreated by placing stacks one atop the other in the same order. For the players' deck, the number of cards in the partitions during the initial setup acts as a guide for the size of each partition mid-way into the game. For instance, in an Easy game (4 epidemics), the initial partitions consists of 13 cards (12 city cards and one epidemic); after two epidemics, if the current player deck has 30 cards, it consists of two partitions with 13 cards (12 city cards and one epidemic), and the top-most partition with 4 cards (and no epidemic, as the epidemic for this partition has already appeared). The forward model shuffles all non-discarded city cards together, then places them into partitions, inserts epidemic cards to partitions that should have one, shuffles the cards in each partition and recreates the player deck by placing partitions in the same order. The forward model randomizes the hidden states of the game in this fashion at the beginning of each gameplay simulation.

\subsubsection{Macro-actions}\label{sec:abstraction_macros}
\begin{figure*}
\centering
\includegraphics[trim=0cm 0cm 0cm 4cm,clip=true,width=0.95\textwidth]{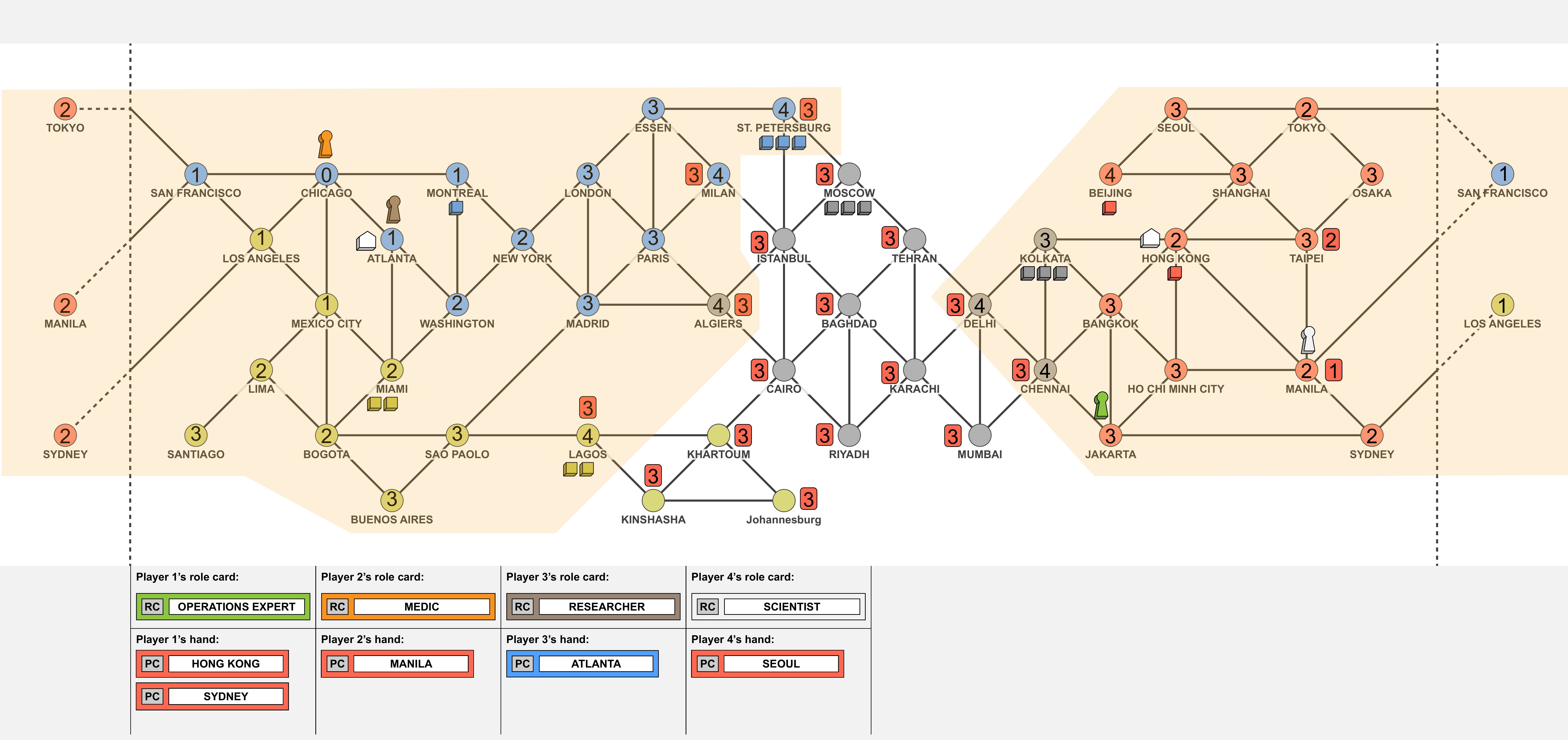}
\caption{Areas accessible to Player~2 (P2) with the different move options (measuring the shortest route), on the game of Fig.~\ref{fig:example}. P2 can move without spending a card to the orange higlighted areas, spending a number of actions shown inside each city. P2 can take a shuttle flight from Atlanta to Hong Kong (both have research stations), increasing reach. P2 can spend the Manila card to travel faster to Manila and Taipei via a direct flight from Chicago. P2 can also travel to Manila via drive/ferry and spend the Manila card there (using the charter flight) to travel anywhere in the word. All cities accessible by spending the red Manila card are shown next to the city with the least actions spent inside a red rectangle.}
\label{fig:example_moves}
\end{figure*}

Each player has four actions per turn, so enumerating all the possible actions would be excessive and could lead to duplicate effort (e.g.  a city can be reached via different routes). To simplify and compress the action state, the concept of macro-actions is introduced. Macro-actions are sequences of actions which actually improve the chance of winning, and combine movement actions to reach a city where that action must be applied. The macro-actions in this implementation are the following:
\begin{itemize}
\item \textbf{Treat disease:} In N actions, reach a city with one or more disease cubes and remove one disease cube (N-1 movement actions, 1 treat disease action).
\item \textbf{Discover cure:} In N actions, reach a city with a research station and discover cure for one (uncured) disease (N-1 movement actions, 1 discover cure action).
\item \textbf{Build research station:} In N actions, reach a city if (a) this player can build a research station there (e.g. by discarding that city card, or for free as the Operations Expert) and (b) this city is at least 4 steps away from another research station, then build a research station (N-1 movement actions, 1 build research station action).  
\item \textbf{Share knowledge (give):} In N actions, reach a city for which the player has that city card and another player would benefit from it. If the other player is already in that city, give the card (N-1 movement actions, 1 share knowledge action). If the other player is not yet in that city or there are no more actions, wait there (N movement actions).
\item \textbf{Share knowledge (take):} In N actions, reach a city where (a) another player is positioned in; (b) the other player has that city card; (c) taking that card is beneficial, then take that card (N-1 movement actions, 1 share knowledge action). If the other player is not yet in that city or there are no more actions, wait there (N movement actions).
\end{itemize}
Each of these macro-actions can include any number of movement actions calculated based on the shortest action sequence (see Fig.~\ref{fig:example_moves}). Eligible movement actions include any drive/ferry action and shuttle flight action (as they do not require spending cards) and any direct flight and charter flight for which the card spent does not reduce the overall chances of curing a disease. The same metric is used to choose cards to discard in case the player has more than 7 cards, and for selecting cards to give or take through the share knowledge macro-actions. For any disease $t$, the ability to cure the disease is measured via $A(t)$ in Eq.~\ref{eq:ability_cure_all} which depends on the best hand across all players (in terms of cards of this type).
\begin{align}
A(t) &=
\begin{cases}
1 &\text{if $t$ cured}\\
max_{p=1{\ldots}P}A_c(p,t) &\text{otherwise}
\end{cases}\label{eq:ability_cure_all}\\
A_c(p,t) &=
\begin{cases}
1, &\text{if ${h(p,t)\geq}C_d(p)$}\\
\frac{h(p,t)}{C_d(p)}, &\text{otherwise}
\end{cases}\label{eq:ability_cure_individual}
\end{align}
\noindent where $P$ is the number of players, $h(p,t)$ is the number of cards of type $t$ in the hand of player $p$ and $C_d(p)$ is the number of cards needed for player $p$ to cure a disease ($C_d=4$ for the Scientist, and $C_d=5$ for every other role).

\subsubsection{State evaluation}\label{sec:abstraction_evaluation}
Taking into account the winning and losing conditions of \emph{Pandemic}, there are several ways to evaluate any given state: \emph{optimistically} in terms of the cards needed to discover every cure, or \emph{pessimistically} in terms of the disease cubes left before the game is lost.
The following state evaluation (fitness) functions are tested in this paper:
\begin{align}
f_{o,d}&=\frac{1}{4}N_d \label{eq:fod}\\
f_{o,A}&=\frac{1}{1.3}\left( \frac{1}{4}\sum_{t=1}^4A(t)+0.3{\cdot}N_d\right) \label{eq:foa}\\
f_{c,a}&=\frac{1}{4}\sum_{t=1}^4\frac{N_{c}(t)}{24} \label{eq:fca}\\
f_{c,m}&=min_{t=1{\ldots}4}\frac{N_{c}(t)}{24} \label{eq:fcm}\\
f_{c,p}&=\prod_{t=1}^4\frac{N_{c}(t)}{24} \label{eq:fcp}\\
f_{b}&=1-\frac{N_b}{8} \label{eq:fb}
\end{align}
\noindent where $N_d$ is the number of cured diseases, $N_{c}(t)$ the number of number of cubes for disease $t$ remaining off the board, and $N_b$ is the number of outbreaks that have occurred so far.

The fitness functions account for cured diseases ($f_{o,d}$) or the general ability to cure diseases ($f_{o,a}$), different ways to calculate disease cubes remaining off the board (average, minimum, or product) and finally the number of outbreaks (as the game ends at 8 outbreaks). All fitness scores are normalized to $[0,1]$ and a high fitness indicates a better game state. Of note is the addition of $0.3{\cdot}N_d$ in Eq.~\eqref{eq:foa} which gives additional pressure if the disease is already cured compared to instances where the disease \emph{can} be cured.

\subsection{Default Policy}\label{sec:agent_default}
Based on the macro-actions defined, a decision-making script was designed based on a hierarchy of macro-actions. This ``default policy'' agent acts as the baseline in our experiments, as the initial individual which the RHEA adjusts through mutations, and also for repairing mutations in RHEA. The default policy enumerates all possible macro-actions of a specific type (based on a predefined order): if there are any macro-actions of this type then a random one of them is chosen and executed, otherwise the next type of macro-actions is enumerated etc. The order was chosen following intuition and experimentation:
\begin{enumerate}
\item Cure disease macro-actions
\item Treat disease macro-actions only for cities with 3 disease cubes of the same type
\item Share knowledge macro-actions (take or give) with immediate effect, otherwise wait in position to share knowledge on another player's turn (take or give)
\item Build research station macro-actions (if there are less than 5 research stations)
\item Treat disease macro-actions only for cities with 2 disease cubes of the same type
\item Treat disease macro-actions only for cities with 1 disease cube of the same type
\item Walk away (i.e. move randomly using all remaining actions left for this player's turn)
\end{enumerate}

Using the game state of Fig.~\ref{fig:example} as an example, we identify the default policy macro-actions on the turn of Player~2 (P2). Since P2 can not discover a cure but can travel to any city if they spend a card (see Fig.~\ref{fig:example_moves}), the default policy will take the 2nd option in the order and choose a random city with 3 disease cubes (St. Petersbourg, Moscow, Kolkata), move there (3 actions in all cases) and spend their last action to remove disease cubes. For the sake of completeness, the next option (3) is to travel to Manila via drive/ferry and give the Manila card to the Scientist: this would increase $A(red)$ of Eq.~\eqref{eq:ability_cure_all} from 2/5 (Player 1's hand) to 2/4 (Player 4's hand). While P2 can build a research station in Manila, that macro-action (4) is unavailable because there is a research station already nearby. Rounding up the other available options, as 5th priority P2 can treat disease at Miami or Lagos (chosen at random), and as 6th priority P2 can treat disease at Montreal, Hong-Kong or Beijing (chosen at random). If no other actions were available, the player would just move to a random location on the board using all 4 actions.

\subsection{Rolling Horizon Evolutionary Algorithm}\label{sec:agent_rolling}
RHEA operates on a horizon of 5 player turns, but operates on the macro-action space. Since macro-actions may take one action (e.g. if the player is already at the right city) but usually take multiple actions, the chromosome for RHEA has a variable length. RHEA is initialized with the macro-actions of the default policy and then applies a 1+1 evolutionary approach, creating a mutation of the current strategy and replacing it if the mutated strategy leads to a higher fitness at the end of the 5 player turns. Evaluation is performed after the player receives new cards at the end of the 5th turn and cities are infected. Note that the forward model is randomized every time an individual is mutated: after each player's macro-actions are simulated, disease cubes are added to cities based on the shuffled infection deck etc.

Based on preliminary parameter tuning, this RHEA implementation applies mutation on every player's turn in the chromosome, choosing one macro-action at random on that player's turn and mutating it. The mutator chooses from an ordered list of macro-actions, where the order is shuffled in each mutation. The candidate macro-actions in their (non-randomized) order are below.
\begin{enumerate}
\item Cure disease macro-actions
\item Treat disease macro-actions only for cities with 3 disease cubes of the same type; if none exist, treat disease macro-actions for cities with 2 disease cubes of the same type; if none exist, treat disease macro-actions for cities with 1 disease cubes of the same type.
\item Share knowledge macro-actions (take or give) with immediate effect, otherwise wait in position to share knowledge on another player's turn (take or give)
\item Build research station macro-actions (if there are less than 5 research stations)
\end{enumerate}
Note that only macro actions that can be completed in this player's turn are considered. For instance, if this is the first macro-action of the player, macro-actions that can be completed in 4 or fewer actions are considered, but if the second or third macro-action of the same player is mutated, the macro-action's duration could be restricted to 1 action. Based on this randomized order, the mutator selects a random macro-action among those in the first set, if there are no such actions then moves to the second set etc. Once a mutated macro-action is selected (and there are still actions remaining for this player after its execution), the default policy is applied again to add macro-actions until the end of this player's turn. This ensures that e.g. if the agents' mutated macro-action moved the player to another location, the agent will not continue with actions that would not be viable. 

Due to the inherent randomness of the forward model, there is a possibility that players' actions (mutated or not) can not be applied. For example, if at the end of the previous turn London was infected and the player's macro-action is to treat disease in London, then in another trial (and shuffled forward model) on the same turn other cities (not London) could be infected and thus the player would have nothing to treat in London. In such cases, the player performs all actions that are viable within the macro-action but ``waste'' actions that are not viable: in the previous example, a player may move from Paris to London (1 action) but not treat disease in London (spending 1 action doing nothing). When evaluating the performance of each individual (the initial default policy and every mutated individual) a number of trials are performed with a shuffled forward model every time, and the state evaluation (see Section~\ref{sec:abstraction_evaluation}) at the end of 5 player turns is averaged to derive the final fitness. 

\section{Experiments}\label{sec:experiments}
In order to assess the performance of the RHEA described in Section \ref{sec:agent}, a number of controlled experiments are carried out in a specific set of game setups (detailed in Section \ref{sec:experiments_testbed}). These controlled experiments explore the impact of state evaluation functions (Section \ref{sec:experiments_evaluation}). Section \ref{sec:experiments_qualitative} compares between the best RHEA agent and the default policy (see Section \ref{sec:agent_default}) which acts as both an initial seed and a baseline for RHEA. Finally, less controlled versions of the same experiments are repeated to assess the robustness of RHEA in more challenging scenarios (Section \ref{sec:experiments_robustness}). All RHEA runs apply a 1+1 evolutionary strategy for 100 generations (population of 1), while the individual's fitness is calculated as the average evaluation of final states in 5 trials, each with a re-randomized forward model.

\subsection{Testbed Setups}\label{sec:experiments_testbed}
\begin{figure}
    \centering
    \includegraphics[width=0.4\textwidth]{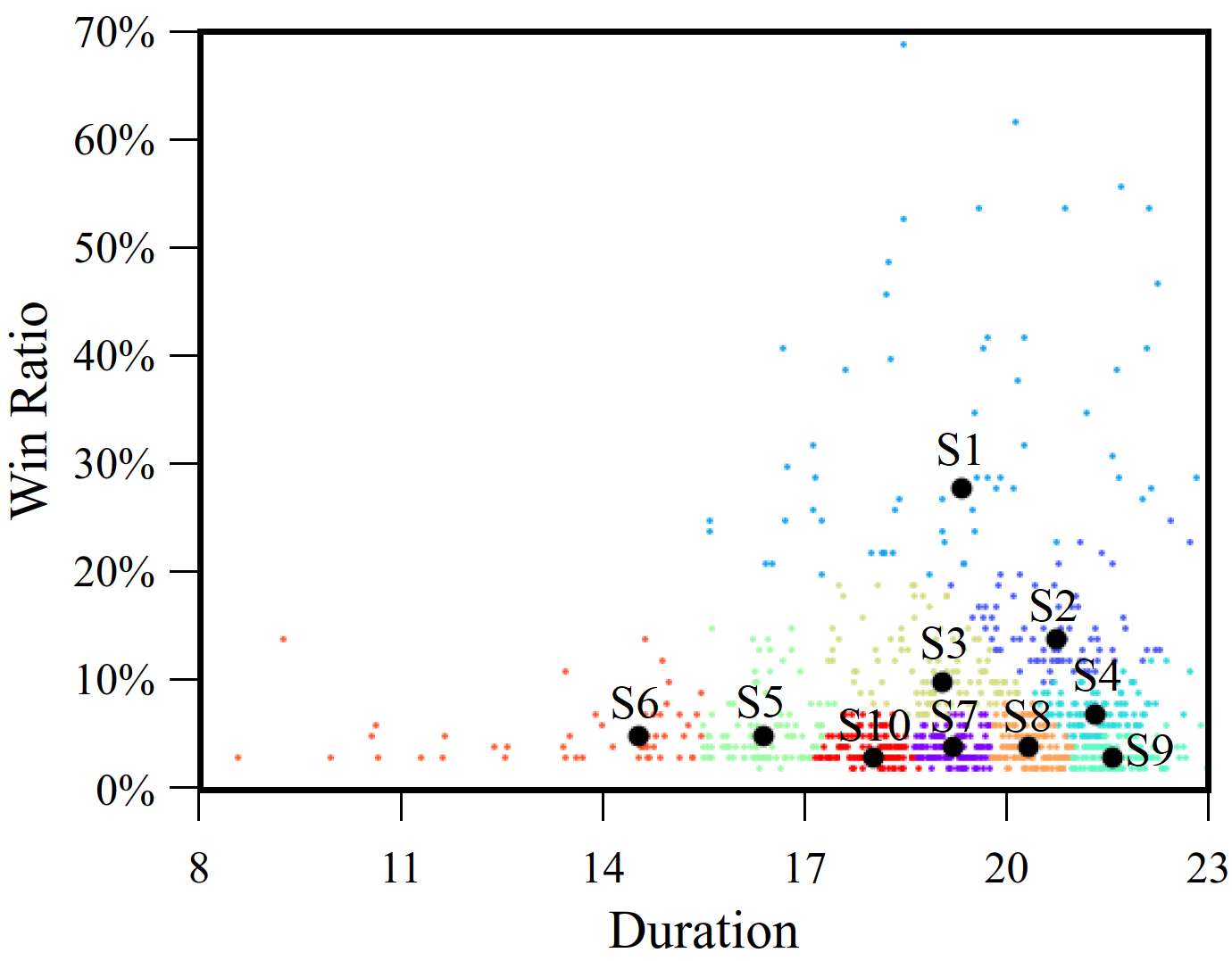}
    \caption{Ten chosen testbed setups (black dots) via $k$-medoids clustering on the 1000 `easiest' setups.}
    \label{fig:clustering}
\end{figure}

The initial game state in \emph{Pandemic} (including the order of cards in the two hidden decks) can greatly affect the game's difficulty. In order to show how the rolling horizon agent can improve the performance of the (scripted yet stochastic) default policy, it is important that many different initial setups are tested. $10^4$ initial setups were created and tested in 100 runs by the default policy until the game was won or lost. The top $10^3$  setups in terms of win ratio were then chosen: those setups had at least a 2\% win ratio and thus did not include setups that were unwinnable. To select a smaller but representative set from these $10^3$ setups, 10 initial setups were selected as the medoids from clustering along the axes of win ratio (naturally between 0 and 1) and duration (normalized based on the maximum game length, i.e. 23 turns). The distribution of the top $10^3$ setups and the 10 medoids are shown in Fig. \ref{fig:clustering}. The average win ratio for the default policy agent for these setups is 8.3\% (ranging from 28\% to 3\% in different cases), and an average game duration of 19 turns (ranging from 13.9 turns to 20.6 turns). Other game metrics point to a variety of strategies favored in each setup: the ratio of share knowledge actions over all actions ranges from 6.1\% to 0.8\%, while ratio of losses due to epidemics ranges from 9\% to 33\%.

Except for Section \ref{sec:experiments_robustness}, all experiments test the 10 chosen setups with the same order of unseen cards, as well as the same four player roles in the same turn order: (1) Operations expert, (2) Medic, (3) Researcher, (4) Scientist. Each setup is played until won or lost for 100 runs. Performance metrics of note is the win ratio in 100 runs, as well as improvement of RH in terms of win ratio over the baseline. The baseline is the default policy (DP) agent, which was also used to select the 10 setups.

\subsection{Impact of State Evaluation}\label{sec:experiments_evaluation}

\subsubsection{Single evaluation}\label{sec:experiments_evaluation_single}
A number of fitnesses are proposed in Section \ref{sec:abstraction_evaluation} for evaluating the state of the game: Eq.~\eqref{eq:fod}-\eqref{eq:foa} are optimistic (taking into account how ``close'' the game is to being won) and Eq.~\eqref{eq:fca}-\eqref{eq:fb} are pessimistic (taking into account how ``far'' the game is to being lost). These fitnesses do not inherently consider whether the game is already won or lost. Variations of each fitness are also tested: Eq.~\eqref{eq:winlose} assigns maximum fitness (1) when the game is won and minimum fitness (0) when the game is lost, while Eq.~\eqref{eq:penalty} rewards winning in the same way but penalizes losing proportionately to the fitness score. The $p(f)$ formula hypothesizes that while losing should always be penalized compared to staying in the game (through a modifier $C_p$), the state of the game when lost can indicate how well the agent could defend against a loss. For all experiments in this paper, $C_p=0.1$.
\begin{align}
w(f)&=
\begin{cases}
1 &\text{if game won}\\
f &\text{if game ongoing}\\
0 &\text{if game lost}
\end{cases}\label{eq:winlose}\\
p(f)&=
\begin{cases}
1 &\text{if game won}\\
f &\text{if game ongoing}\\
C_p{\cdot}f &\text{if game lost}
\end{cases}\label{eq:penalty}
\end{align}

\begin{figure}
\subfloat[State evaluation as single fitness]{\includegraphics[trim=3mm 5mm 4mm 5mm, clip,height=0.18\textwidth]{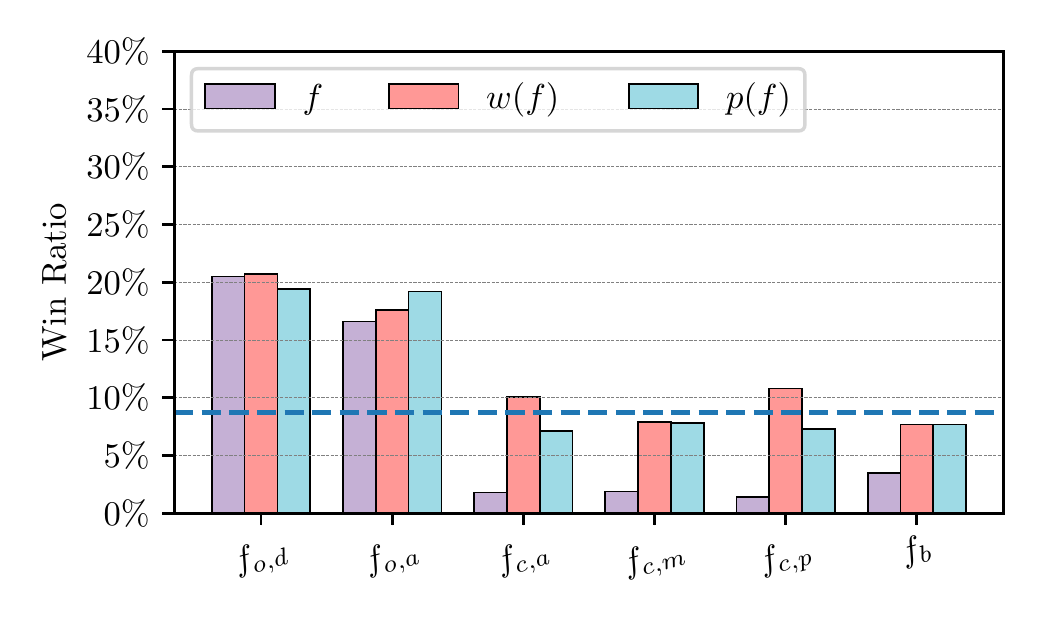}
\label{fig:experiments_evaluation_single} 
}\\
\subfloat[State evaluation as average fitness of the two fitness scores]{
\includegraphics[trim=3mm 5mm 4mm 5mm, clip,height=0.18\textwidth]{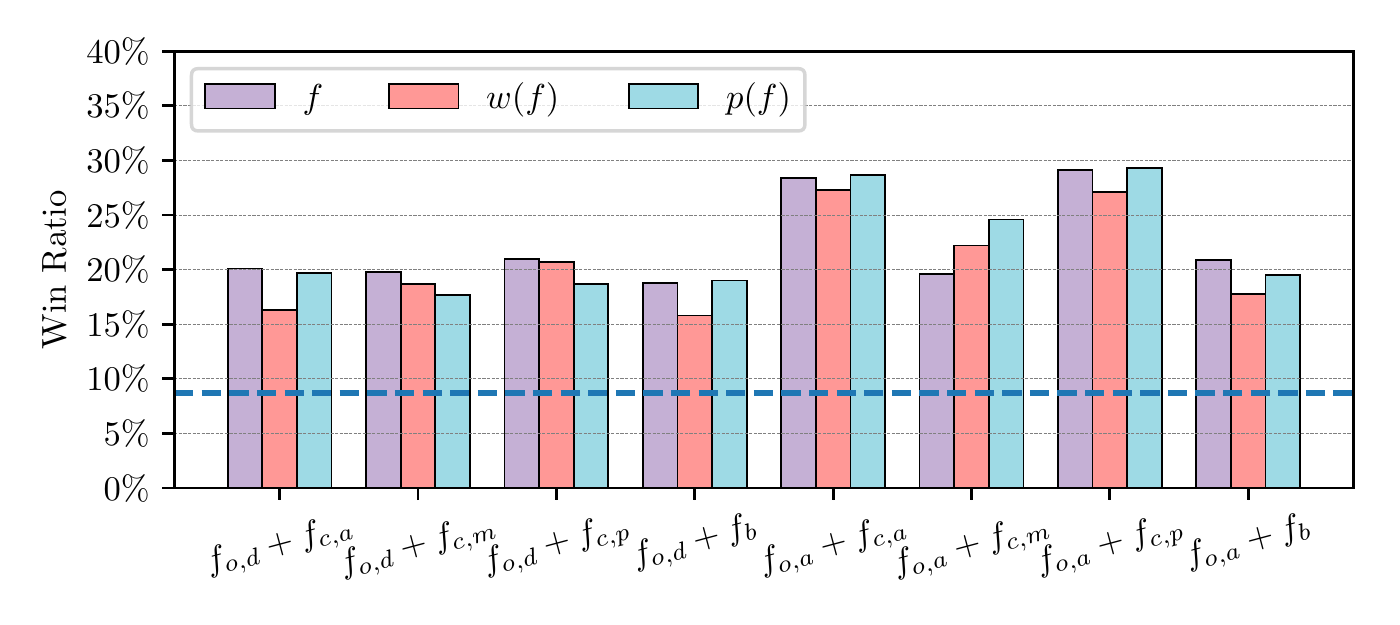}
\label{fig:experiments_evaluation_aggregated}
}
\caption{Average win ratio for the 10 test setups, using different state evaluations (with or without win/loss conditions). The dotted line is the win ratio of the default policy.}\label{fig:experiments_evaluation} 
\end{figure}

This experiment tests each fitness of Section \ref{sec:abstraction_evaluation} in its 3 variants: the average win ratio in the ten testbed setups (from 100 trials in each setup) are reported in Fig.~\ref{fig:experiments_evaluation_single}. An important observation is that pessimistic evaluations on their own perform much worse than the default policy, or comparably when winning and losing conditions are accounted for. Generally, the $w(f)$ variant performs better than the default fitness, while only for $f_{o,a}$ the penalty seems to have a positive effect. The optimistic fitnesses manage to steer the agent towards winning the game more often: the best improvement over the default policy (averaged across the ten setups) is 120\% with $w(f_{o,d})$. In terms of other differences between the agents, optimistic agents generally tend to play shorter games and lose much faster than pessimistic agents. Indicatively, $f_{o,d}$ is the fastest to lose, with lost games' average duration at 14.3 player turns, versus $f_{c,p}$ which is the slowest to lose (21.3 turns). Unsurprisingly, pessimistic agents who prioritize keeping disease cubes off the board rarely lose due to outbreaks or insufficient cubes: indicatively, $f_{d,p}$ loses due to epidemics in 22\% of lost games and due to disease cubes in 6.3\% of lost games, compared to 56\% and 24\% respectively for $f_{o,a}$. Finally, optimistic agents tend to use the share knowledge action more often than the DP agent while the opposite is true for the pessimistic agents. Pessimistic agents tend to use the treat disease action more often than the DP agent, while the opposite is true for optimistic agents. These differences in actions favored are less pronounced when the fitness is conditionally applied as $w(f)$ or $p(f)$.

\subsubsection{Combined evaluation}\label{sec:experiments_evaluation_aggregated}

While fitness functions measuring how close players are to winning seem to perform well, optimistic RHEA agents underestimate losing conditions and tend to lose quickly. The hypothesis is that combining optimistic and pessimistic fitnesses could allow agents to account for both opportunities and dangers in their final state. For the sake of this experiment, two fitness scores are averaged (one optimistic, one pessimistic) and applied either on their own or conditionally via Eq.~\eqref{eq:winlose} and Eq.~\eqref{eq:penalty}. 

The average win ratio in 10 setups (from 100 trials per setup), for different combinations of state evaluations are reported in Fig.~\ref{fig:experiments_evaluation_aggregated}. 
Interestingly, a trend is reversed compared to Fig.~\ref{fig:experiments_evaluation_single} in that the naive aggregated state evaluation often performs better than the conditional variants, especially $w(f)$. While $f_{o,d}$ performed better on average than $f_{o,a}$ when applied alone, in this case fitnesses that combine $f_{o,a}$ perform much better. While differences are quite small among the most well-performing agents, the best agent is $p(\tfrac{f_{o,a}+f_{c,m}}{2})$ with an average win ratio of 29.3\%, i.e. an average improvement of 302\% over the DP agent (calculated per setup). This fitness will be used in the next experiments for the RHEA agent.

\begin{figure}
\includegraphics[trim=3mm 5mm 3mm 3mm, clip,width=0.4\textwidth]{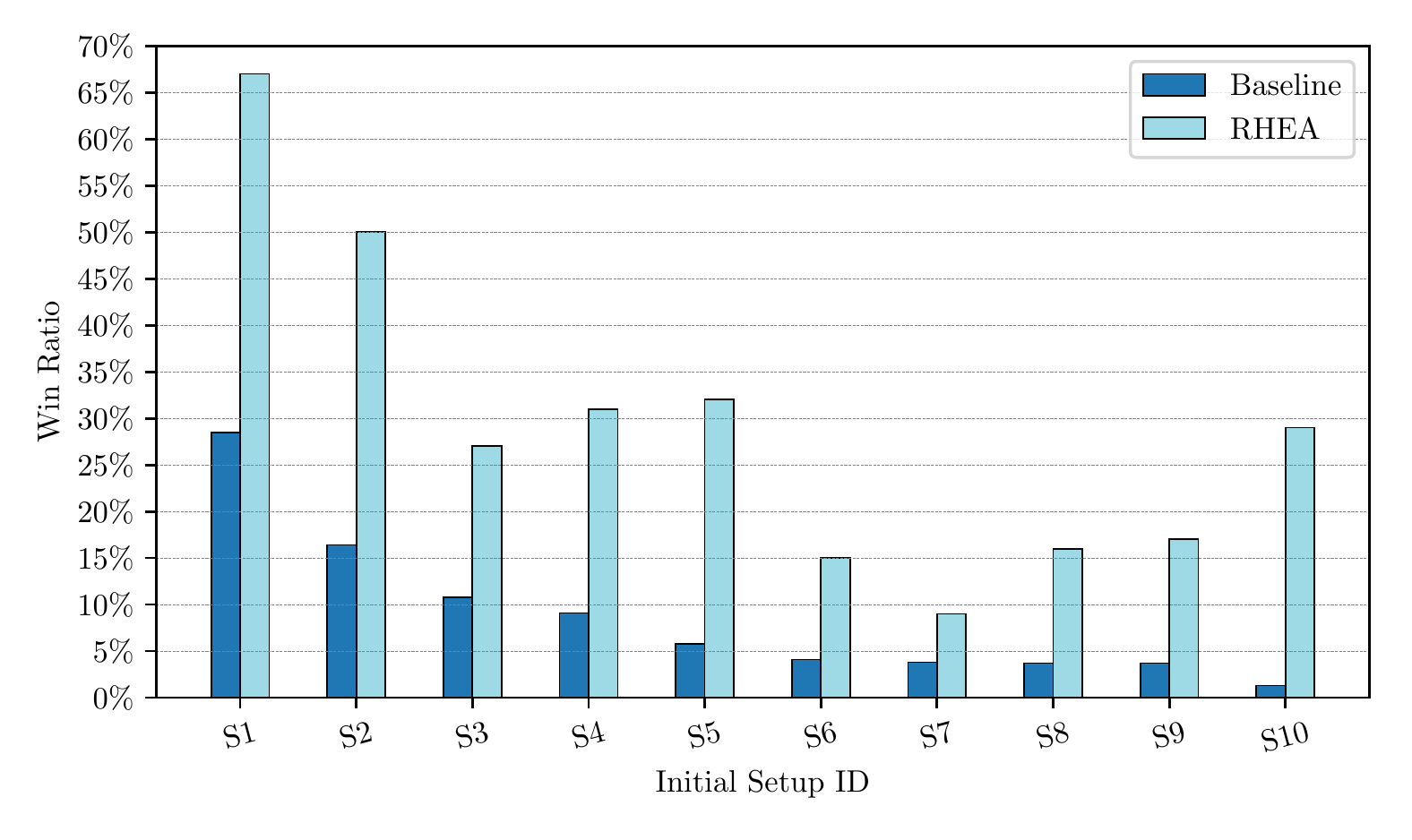}
\caption{The win ratio of the default policy and the best RH agent per initial setup.}\label{fig:experiments_perlevel_win} 
\end{figure}
\begin{figure}
\includegraphics[trim=3mm 5mm 3mm 3mm, clip,width=0.45\textwidth]{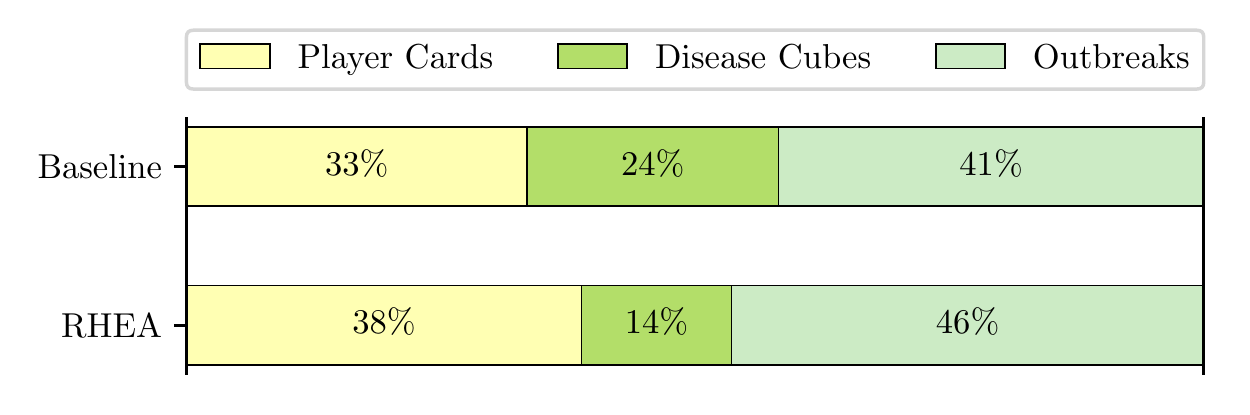}
\caption{The ratio of losing conditions triggered per agent.}\label{fig:experiments_ratio_lost} 
\end{figure}
\begin{figure}
\includegraphics[trim=3mm 5mm 3mm 3mm, clip,width=0.45\textwidth]{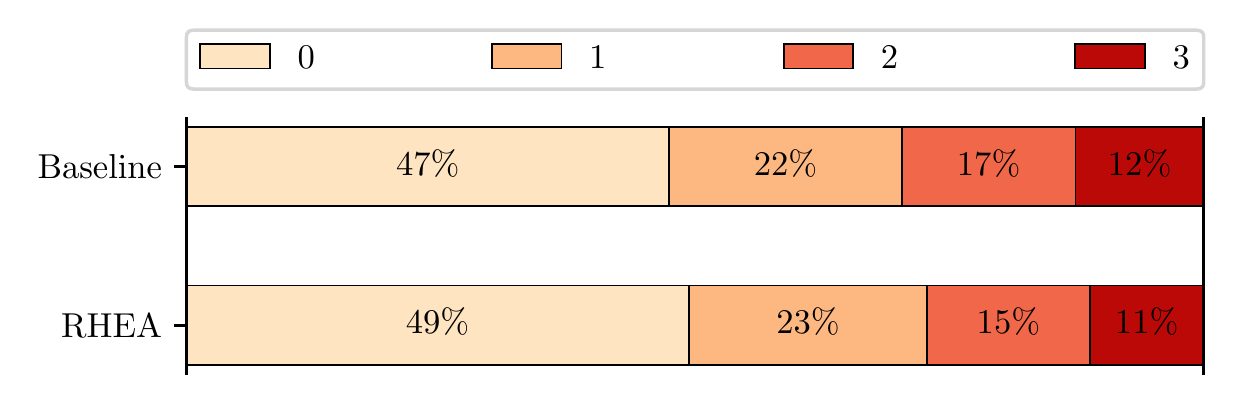}
\caption{Ratio of cities with the listed number of cubes of the same type at the end-game, per agent.}\label{fig:experiments_ratio_infected_cities} 
\end{figure}

\subsection{Analysis of RHEA Strategies}\label{sec:experiments_qualitative}

With the best RHEA agent discovered through the exploration of Section \ref{sec:experiments_evaluation}, it is important to identify the differences in performance and behavior of RHEA with the baseline DP used to initialize the macro-action sequence. This section explores the impact of the ten setups, as well as the changes in decision-making and end-game statistics, comparing the best RHEA agent and the DP agent.

Figure \ref{fig:experiments_perlevel_win} shows the win ratio of the two agents in the ten  setups tested in this paper. The setups are sorted based on the behavior of the default policy. While the RHEA agent improves performance over the baseline in all cases (average improvement over the baseline is 302\%), it performs best in very difficult setups such as S10 (with a win ratio over 22 times that of the baseline). In some setups such as the ``easy'' S1 (baseline win ratio of 29\%) or the ``difficult'' S7 (baseline win ratio of 3.8\%), the improvements for RHEA are not as pronounced (135\% improvement in S1 and 137\% in S7).

Figure \ref{fig:experiments_ratio_lost} shows the ratio of each losing condition triggered. It is evident that the RHEA agent lost more often because the maximum turn limit was reached (i.e. due to no more player cards available to draw from). While RHEA was also better at managing the disease cubes on the board (losing far less often due to insufficient disease cubes), it did not seem able to avoid outbreaks. Averaging across all 10 setups, the RHEA agent achieved a 9\% drop in the number of outbreaks compared to the baseline, although in some setups the difference was more pronounced (e.g. 21\% fewer outbreaks for S3). The improved strategy of RHEA in handling disease cubes is verified in Fig.~\ref{fig:experiments_ratio_infected_cities} which shows the ratio of cities with disease cubes of the same type at the end-game. Evidently, RHEA can keep the number of cities with three disease cubes slightly lower, thus lowering the chance of an outbreak.
On the other hand, the ratio of cities with one disease cube does not change as they rarely trigger an outbreak.

\begin{figure}
\includegraphics[trim=3mm 5mm 3mm 3mm, clip,width=0.45\textwidth]{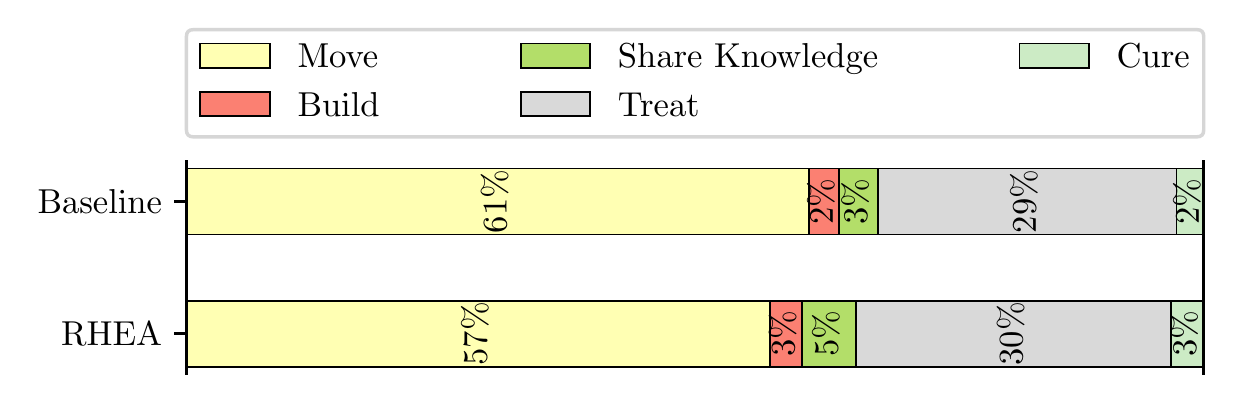}
\caption{The ratio of actions of each type, taken from every player, per agent.}\label{fig:experiments_ratio_actions} 
\end{figure}

Figure~\ref{fig:experiments_ratio_actions} shows the ratio of each type of action in the playthroughs of the best RH agent and the baseline. While both agents primarily spend most of their actions moving around the board, RHEA spends fewer actions doing so. Interestingly, RHEA also spends slightly fewer actions treating disease despite the fact that games end with fewer cubes on the board and fewer outbreaks (see Fig.~\ref{fig:experiments_ratio_lost}). RHEA spends only marginally more actions building research stations; the number of research stations at the end-game is only 11\% higher for RHEA compared to the DP agent. Clearly, the primary difference in strategy is that the RHEA agent favors the share knowledge action, choosing it almost twice as often (96\% increase). In some games this was even more pronounced, e.g. in S7 and S5 the share knowledge set of actions was chosen approximately four times as often (301\% increase and 275\% increase respectively).

\subsection{Testing Robustness}\label{sec:experiments_robustness}
In all the experiments so far, the same set of 10 setups were tested, with a total of four epidemics in the player deck and with a preset player order. Due to the way in which the initial game states were selected (clustering based on the baseline performance), the experiments were highly controlled as the order of the hidden decks was always the same and the difficulty of the game was Easy (based on the rules of \emph{Pandemic}). Having established the differences in agents' performance in controlled experiments, it is important to also test the robustness of the RHEA agents' performance when the order of play, the hidden decks, and the number of epidemics changes. This Section performs a number of experiments on the same 10 setups but (a) randomizing the order of players' roles in every trial ($P_{rand}$), (b) shuffling the infection and player decks ($D_{rand}$) after the initial cities are infected and starting player cards are given, and (c) changing the number of epidemics in the player deck from 4 (Easy difficulty), to 5 (Medium difficulty), and 6 (Hard difficulty).

\begin{figure}
\includegraphics[trim=3mm 4.5mm 3mm 4mm, clip,height=0.18\textwidth]{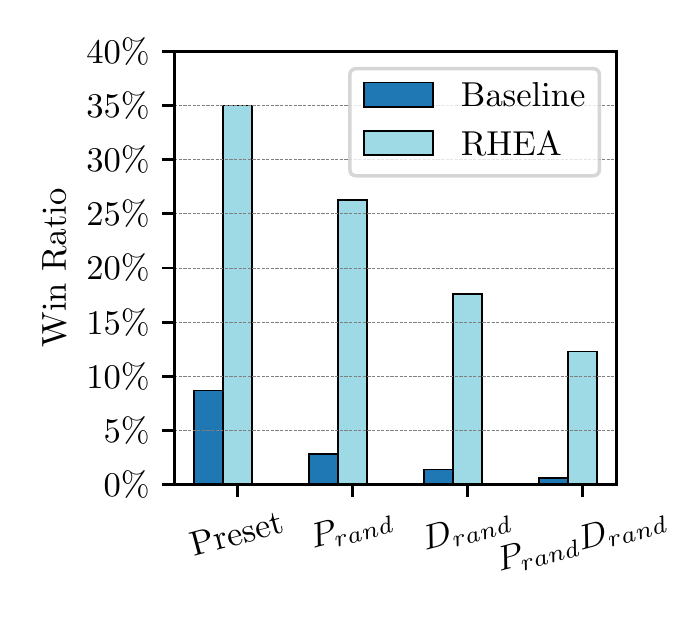}
\caption{Impact of different randomizations on the initial setups, on Easy difficulty and 4 players.}\label{fig:robust_preset} 
\end{figure}

Fig.~\ref{fig:robust_preset} shows how the win ratio drops for both agents (baseline and RHEA) when the player order and/or the hidden decks are randomized. Interestingly, a simple reordering of player roles seems to severely affect the baseline agent (68\% drop in win rate), while the RHEA agent is less sensitive to this (25\% drop in win rate). Unsurprisingly, the initial game states were likely selected because the cards in the hidden decks resulted in fairly easy game progressions; when the hidden decks are randomized the drop is substantial for both agents (84\% drop for DP, 50\% drop for RHEA). Clearly when both hidden decks and player order is randomized the task becomes more challenging, although the RHEA agent still manages to win $12\%$ of the games. Indeed, as the games become more challenging due to the randomizations, RHEA outperforms the baseline by a larger margin (winning up to 20 times more often when both hidden decks and player order is randomized).

\begin{figure}
\includegraphics[trim=3mm 4.5mm 3mm 4mm, clip,height=0.18\textwidth]{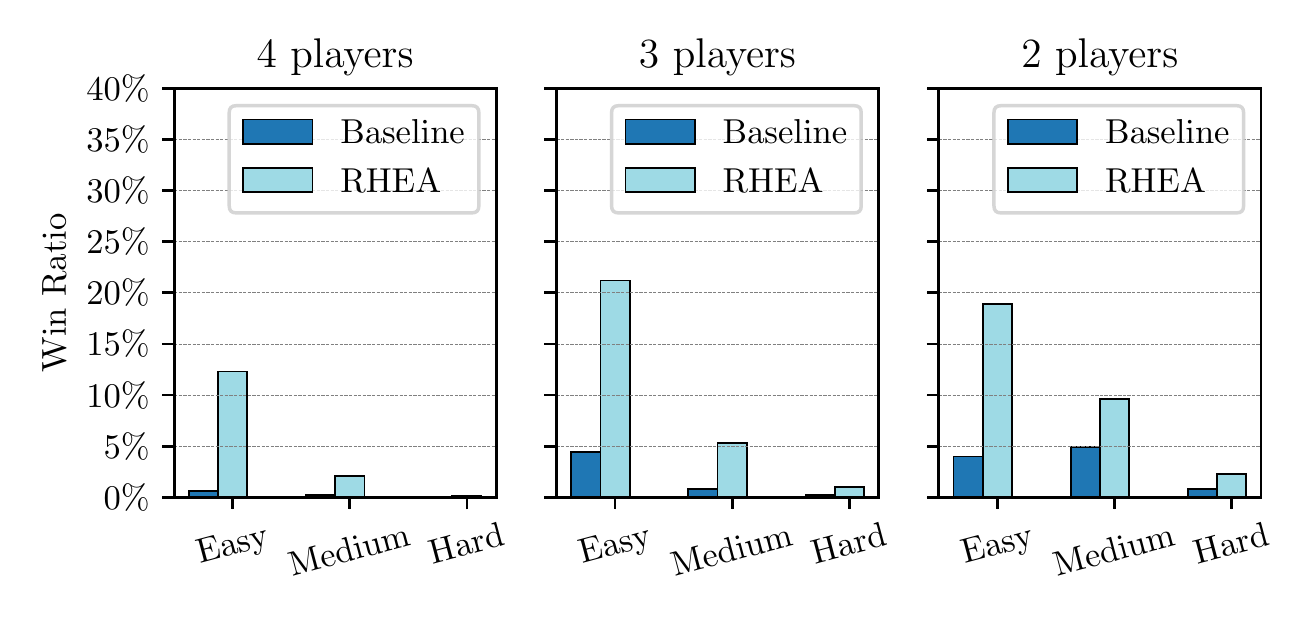}
\caption{Impact of different players when the number of epidemics changes. Both player order and hidden decks are randomized ($P_{rand}D_{rand}$) in these experiments. Results are averaged from 100 trials in each of the 10 testbed setups.}\label{fig:robust_difficulties_players} 
\end{figure}

Other factors that affect the challenge is the number of epidemics and the number of players. As the number of epidemics increases, the same cities become infected more often and the likelihood of outbreaks increases. On the other hand, if \emph{Pandemic} is played by fewer players then it is easier to plan ahead and coordinate as the game state changes less between a player's consecutive turns. The impact of both the number of players and the number of epidemics is shown in Fig.~\ref{fig:robust_difficulties_players}; in these experiments, both player order and hidden decks are randomized. Specifically, the player deck is split into stacks of the same size and the epidemics are added to each (as normal), while in games with fewer players the roles are randomly selected in each trial among the four tested in 4-player games (Operations expert, Scientist, Researcher, Medic). Results from Fig.~\ref{fig:robust_difficulties_players} show that, unsurprisingly, games become easier with fewer players even when more epidemic cards are added. While for 3 players an Easy \emph{Pandemic} game is less challenging for both RHEA and DP agent than a 2-player game (likely due to better synergies between different roles), the reverse is true for Medium and Hard games. Admittedly, in Hard difficulties RHEA also suffers (with win ratios below 3\% even with two players). However, RHEA can win games in which the baseline never finds a winning strategy. In Hard difficulties, DP manages to win any of the 100 trials in 5 out of 10 setups when playing with 2 players, and in 2 out of 10 setups when playing with 3 players; RHEA manages to win at least once in 100 trials in 9 out of 10 setups and in 7 out of 10 setups when playing with 2 and 3 players respectively.

\section{Discussion}\label{sec:discussion}
Experiments in this paper have illustrated that a rolling horizon evolutionary algorithm can enhance the performance of the well-designed baseline agent for playing \emph{Pandemic}, winning four times as often in the controlled testbeds prepared for this paper. Moreover, in more difficult conditions the RHEA is able to discover more winning strategies. It is hypothesized that the improved performance of RHEA is due to the fact that it can anticipate better the upcoming challenges through multiple simulations of the forward model, but also because it can better adapt macro-action selection to the strengths of each player role (e.g. prioritize share knowledge macro-actions for the researcher and treat disease for the medic).

It should be noted that the strictly defined baseline agent, which also acts as seed and repair mechanism for the RHEA, is also a weakness of the proposed method. First of all, the RHEA has only a limited degree of freedom in terms of its available strategies, and those are further arbitrarily limited in terms of e.g. proximity constraints on where research stations can be built. On the other hand, preliminary experiments with fewer constraints on actions (e.g. mutation being able to choose actions that would not be completed in this player's turn) exhibited poor performance. RHEA would need far more computational resources to perform well if the number of options is not carefully controlled. Other rules, such as when players can share knowledge (only when $A(t)$ would increase) are beyond the control of the RHEA. This likely explains why share knowledge actions were rarely chosen, and ultimately why higher difficulty games were rarely won. Finally, the experiments were performed on testbed setups which could be solved by the DP agent: this could bias the findings by testing games that e.g. did not require as much knowledge sharing in order to be won. The experiments in Section \ref{sec:experiments_robustness} showed how the baseline rule-based agent underperforms when the hidden states are randomized, while RHEA is better able to adapt to less favorable test conditions. 

It is important to note that current experiments exclude certain player roles and all special event cards, which makes the game far more challenging. Specifically, event cards add more player turns (as they are added to the player deck) and also allow for emergency actions outside the players' 4 actions per turn. However, dealing with actions that can be taken on another player's turn or after drawing player cards (via special events), or actions that move other players (via the Dispatcher player role) would highly complicate the representation of the RHEA chromosome.

Experiments in Section \ref{sec:experiments_evaluation} showed how different lenses of assessing the game state can lead to different strategies and performance: pessimistic evaluations which tried to avoid a premature loss favored treating diseases while optimistic evaluations which tried to get closer to a win favored sharing knowledge. Experiments in Section \ref{sec:experiments_evaluation_aggregated} showed that a simple aggregation of optimistic and pessimistic evaluations can lead to a much improved performance. Other combinations of state evaluations likely performed worse due to the relative imbalance between the two metrics combined. Adjusting the weight of pessimistic versus optimistic evaluation will likely improve performance further, via e.g. a weighted sum of two or more state evaluations of Eq.~\eqref{eq:fod}-\eqref{eq:fb}. A multi-objective approach for RHEA could also better handle the tradeoff between the likely conflicting state evaluations. However, preliminary experiments with a simple algorithm based on NSGA-II \citep{Deb2002nsga2} where pessimistic and optimistic state evaluations were combined yielded poor results, likely requiring more computational resources (e.g. generations or population size) to reach peak performance.

Further exploration in the vein of collaborative board game play for Artificial Intelligence could target more complex games. As noted in Section \ref{sec:introduction}, games such as \emph{Arkham Horror}, \emph{Robinson Crusoe}, or \emph{Zombicide} (CMON, 2012) have similar design patterns with \emph{Pandemic} but complicate the game state substantially via player inventories (with items that may further modify the optimal or allowed actions that players can take) and further randomness as players roll dice for combat and other actions. Further research in abstracting the game state, likely via machine-learned forward models \citep{Lucas2019forwardmodellearning} based on a corpus of playthroughs (or parts of a playthrough, such as a combat sequence), would be necessary in order to playout such a game. Another avenue for exploration would be having both AI and human players collaborating in the same game, e.g. where one human player can play a game intended for 4 players by offloading the other roles to the AI. In this case, the primary challenges for an AI would be (a) anticipating the most likely actions of the player in its forward model, and (b) explaining and guiding the player on the strategy the AI players wish to follow. For the former challenge, the AI will need to model the human co-players, either based on their past actions \citep{Yannakakis2013PlayerM} (in a data-driven manner) or as procedural personas \citep{holmgard2014evolvingpersonas} of board game players (e.g. the altruist or the egoist). For the latter challenge, research in explainable AI \citep{zhu2018explainable} can be carried out regarding suggestions of action and cost/benefit visualizations so that human players are convinced of the best course of action, or suggest their own strategies for the AI to follow.

\section{Conclusion}\label{sec:conclusion}
This paper has highlighted the challenges and opportunities that collaborative board games such as \emph{Pandemic} pose to AI agent control, due to the complex task of controlling the stochasticity of the environment and anticipating immediate and long-term dangers. Moreover, the different roles that players take in a game of \emph{Pandemic} and the high branching factor of the game's original actions necessitate several innovations in the design of an AI controller that can strategize all players' actions efficiently. The RHEA tested in this paper seems to perform competently in many different environments, difficulty levels, and number of players. However, further research in \emph{Pandemic} game play can shed important light on AI agent control in the underexplored domain of collaborative board game play.

\begin{acks}
The authors would like to thank Alberto Tonda for early discussions and formalizations of the problem of \emph{Pandemic} agent control. This project has received funding from the European Union’s Horizon 2020 programme under grant agreement No 787476. 

The code for the \emph{Pandemic} framework and the reported AI agents can be found at \url{https://github.com/konsfik/Pandemic-AI-Framework}.
\end{acks}

\bibliographystyle{ACM-Reference-Format}
\bibliography{pandemic} 

\end{document}